# Inroads Toward Robot-Assisted Internal Fixation of Bone Fractures Using a Bendable Medical Screw and the Curved Drilling Technique

Farshid Alambeigi[1], *Member, IEEE*, Mahsan Bakhtiarinejad[1], Armina Azizi[2], Rachel Hegeman[1,4], Iulian Iordachita[1], *Senior Member, IEEE*, Harpal Khanuja[3], and Mehran Armand[1,4], *Member, IEEE*

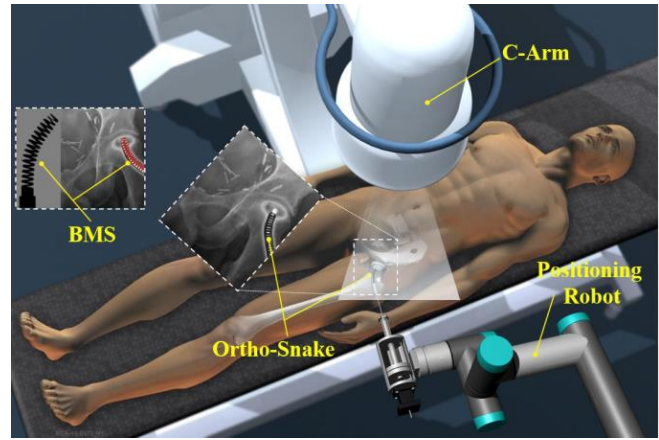

Fig. 1. Conceptual illustration of the proposed robotic workstation for internal fixation of bone fractures using curved drilling technique. It comprises of a positioning robot, a continuum manipulator (i.e. ortho-snake), proper flexible cutting tools, and a bendable medical screw for fixation of the fracture.

*Abstract*— Internal fixation is a common orthopedic procedure in which a rigid screw is used to fix fragments of a fractured bone together and expedite the healing process. However, the rigidity of the screw, geometry of the fractured anatomy (e.g. femur and pelvis), and patient's age can cause an array of complications during screw placement, such as improper fracture healing due to misalignment of the bone fragments, lengthy procedure time and subsequently high radiation exposure. To address these issues, we propose a minimally invasive robot-assisted procedure comprising of a continuum robot, called *ortho-snake*, together with a novel bendable medical screw (BMS) for fixating the fractures. We describe the implementation of a curved drilling technique and focus on the design, manufacturing, and evaluation of a novel BMS, which can passively morph into the drilled curved tunnels with various curvatures. We evaluate the performance and efficacy of the proposed BMS using both finite element simulations as well as experiments conducted on synthetic bone samples.

## I. INTRODUCTION

Recent literature documents a range of successfully performed robotically-assisted orthopedics procedures (RAOPs) including hip replacement [1], spine surgery [2], and fracture treatment [3]. Despite promising results achieved by implementing robot-assisted technologies in orthopedics, most of RAOPs are still utilizing rigid and non-dexterous instruments and methods [3]. Considering the current advancements in technology and fabrication methods, novel assistive devices for Minimally Invasive Surgery (MIS) may provide better patient outcomes and reduce complexities associated with MIS procedures.

Internal fixation is a common orthopedic procedure in which typically a rigid cannulated screw is used to fix fragments of a fractured bone together and expedite the healing process [4]. However, the rigidity of the screw, geometry of the fractured anatomy (e.g. bones found in shoulders, wrists, pelvis and equivalents thereof), and patient's age can cause an array of complications during the screw's placement. For instance, internal fixation of the femoral head with traditional rigid screws is a common and costly procedure done on elderly patients which has a low success rate [5]. This is due to the fact that the screw is typically implanted into osteoporotic bone and, therefore, cannot provide sufficient hold during the healing process, leading to a fixation failure. Further, fracture fixation using existing screws in bones with complex anatomy (e.g. pelvis) may become very cumbersome and require multiple trials [6], [7]. These trials increase the surgery time and radiation exposure to patients and clinicians.

To address limitations in maneuverability and dexterity of conventional medical instruments, various flexible medical devices, techniques, and lately continuum robots have been developed to enhance clinicians' accessibility in confined environments. However, most of these technologies have been developed for surgeries interacting with soft tissues (e.g. [8], [9], [10], [11]). With the goal of enhancing the dexterity and accuracy, our group is focusing on developing a surgical robotic system for internal fixation of bone fractures using a custom designed continuum robot called *"ortho-snake"* (Fig. 1). Continuum manipulators are commonly designed to interact with soft tissues. Orhto-snake, thanks to its structural stability, can robustly interact with hard tissues and bear high external loads during bone milling [12] and drilling [13].

In this paper, we describe the implementation of a curved

*Research supported by NIH/NIBIB grant R01EB016703.

[1]Farshid Alambeigi, Mahsan Bakhtiarinejad, Rachel Hegeman, Iulian Iordachita, and Mehran Armand are with Laboratory for Computational Sensing and Robotics, Johns Hopkins University, Baltimore, MD, USA, 21218. Email:{falambe1,mbnejad,rhegema1,iordachita,marmand2}@jhu.edu

[2]A. Azizi is with the Department of Plastic and Reconstructive Surgery, Johns Hopkins Medical School, Baltimore, MD 21205 USA (e-mail:aazizi3@jhmi.edu).

[3]H. Khanuja is with the Department of Orthopedic Surgery, Johns Hopkins Medical School, Baltimore, MD 21205 USA (e-mail:hkhanuj1@jhmi.edu).

[4]Rachel Hegeman and Mehran Armand are also with Johns Hopkins University Applied Physics Laboratory, Laurel, MD, USA, 20723. Email: {rachel.hegeman, mehran.armand}@jhuapl.edu.

drilling technique with the ortho-snake for internal fixation of fractures during MIS. With this novel steerable drilling technique, clinicians have more freedom when selecting the entry point into the bone as compared to common straight drilling techniques. Further, they have the ability to steer the drill to make a tunnel that may provide enough intraoperative interfragmentary compression after fixation by a proper screw [13]. In this paper, we investigate the feasibility of using curved drilling technique together with a novel bendable medical screw (BMS). We discuss the design requirements, finite element analysis and manufacturing procedure of the BMS. Moreover, performance of the manufactured BMS is tested on curved tunnels drilled with the robotic system utilizing the ortho-snake (Fig. 1).

## II. INTERNAL FIXATION WITH A BENDABLE SCREW

Medical screws are typically used for internal fixation of bones and placed across the fractures to bring the surfaces of the fracture in close proximity and compression [14], [4]. Conventional screws are solid devices consisting of a head, a straight shaft, which may be fully or partially threaded and cannulated, and a tip, which may or may not have a self-tapping design. The main challenge of internal fixation using traditional straight screws is placement of a rigid device into bones that may have curved anatomy e.g. femoral head, pelvis, and bones in shoulders, wrist and equivalent thereof [5], [7]. Of note, failure to achieve an accurate placement can result in improper bone reunion due to misalignment of the bone fragments.

To provide more dexterity and flexibility for clinicians during screw placement, in this paper, we introduce a robotically-assisted orthopedics procedure for internal fixation with a passive BMS. In this approach, first, the broken bone is reduced or put back into place and based on the clinician's decision, a curved tunnel is initially drilled utilizing a robotic system. Then, to fixate the fractured bone, a BMS is placed inside the tunnel. In the remainder of this section, we outline the system design requirements and the steps taken to satisfy them:

(i) *The robotic system should drill a straight or curved tunnel.* To satisfy this condition, we mainly rely on the robotic system described in Section IV and the curved drilling technique performed by the ortho-snake and the appropriate flexible drill bit as shown in Fig.5a. The ortho-snake is made of nitinol tubes (with outer diameter of 6 mm and inner diameter of 4 mm) with notches along its length, which constrains its bending to a single plane via two actuating tendons [15]. The drill bit with 8.5 mm outer diameter was fabricated from a ball-end carbide end mill (8878A18, McMaster-Carr) with two flutes and helix angle of 30°. The performance and capability of this robotic system have been successfully evaluated on both simulated bones and cadaveric specimens [13].

(ii) *The BMS should passively and safely follow the curved tunnel.* Unlike rigid screws the BMS axis of rotation is not straight and follows the drilled curved tunnel. Also the twist axis of the tightening/loosening torque applying on the BMS head is not aligned with the variable twist axis and the translational axis of its shaft. Hence, the BMS needs to freely bend in different planes in order to transfer the torque and translate the BMS along the curved tunnel. This freedom in bending motion during applying the torque on the BMS head can be achieved by using orthogonal notches (as marked by numbers 3 and 4 in Fig. 2) along the shaft of the BMS. Of note, geometry of these notches (i.e. their type, depth, and pitch) determines the flexibility of the BMS when interacting with the cancellous bone, while translating through the tunnel. Also, to improve passive steerability and guarantee a safe interaction between the tip of the screw and the bone, a round steering head is used for the screw (as marked by number 6 in Fig. 2).

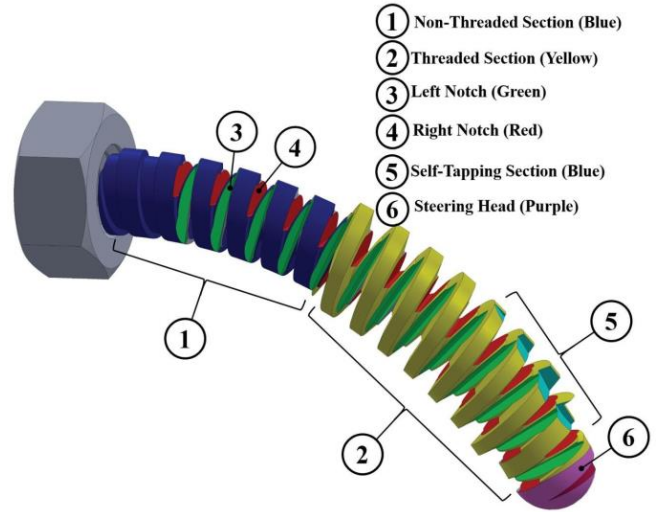

Fig. 2. Conceptual design of a BMS showing the considered features to address the design requirements of a robot-assisted internal fixation procedure.

(iii) *The BMS should tap the surface of the curved tunnel during its placement into the bone (i.e. self-tapping capability).* As shown in Fig. 2, this feature is created by designing a sharp edge, similar to those on a tapping tool, on the surface of the threads located at the tip of the screw. This sharp section (as marked by number 5 in Fig. 2) carves precise groves on the surface of the cancellous bone to pave the path for the rest of BMS threads that follow this section.

(iv) *The BMS should provide a stable fixation for the fracture site.* Parameters affecting the holding strength of the conventional rigid screws include: screw geometry (i.e. type, depth, and number of threads), outer diameter, core diameter, and pitch of the BMS (as shown by numbers 1 and 2 in Fig. 2), the screw material, and the ultimate shear strength and density of the bone [14]. Of note, for the BMS, the screw geometry not only affects the flexibility and shape-adaptability of the screw but it also plays an important role in its internal fixation efficacy and strength. There is a trade-off between the screw flexibility and strength, which needs to be optimized prior to fabrication based on the fixation procedure and location of the fracture.

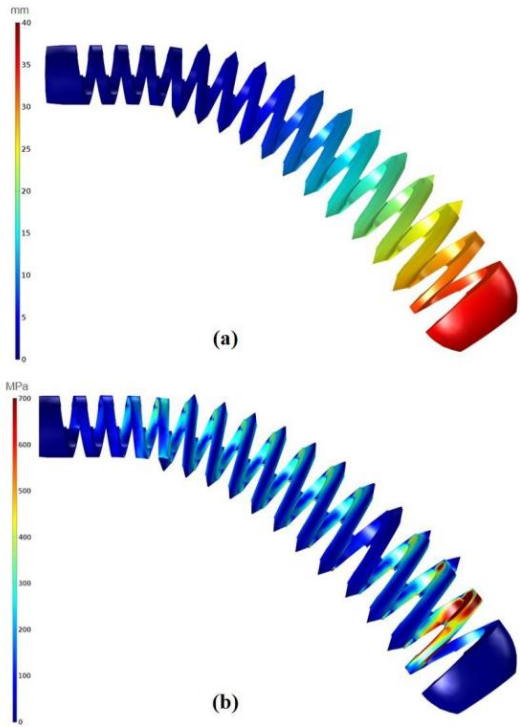

Fig. 3. The performed FE analysis to evaluate the shape adaptability of the designed BMS. In this simulation, we assume the BMS has been inserted into the quadratic shape curved tunnel. (a) considered quadratic displacement boundary condition along the BMS with maximum displacement of 40 mm showing the imposed geometry of the tunnel on the BMS; (b) distribution of the stress along the BMS considering the applied boundary condition.

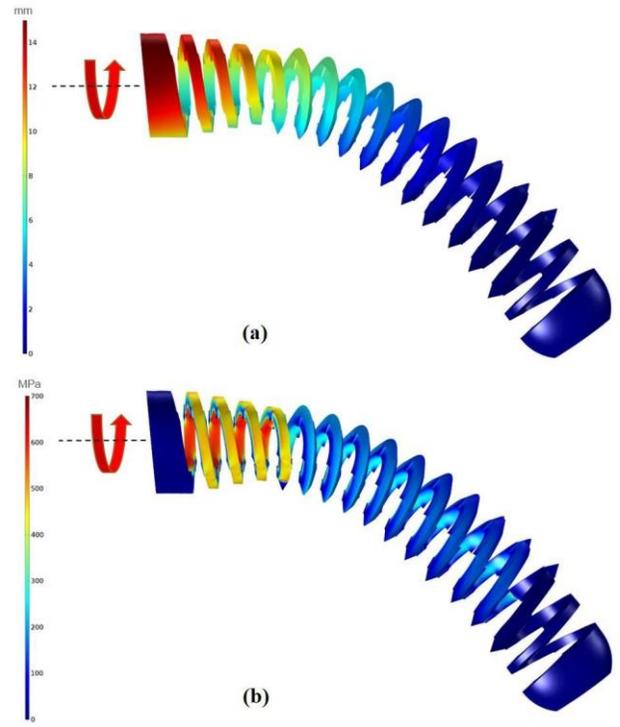

Fig. 4. The performed FE analysis to evaluate the tapping capability of the designed BMS. In this simulation, we assume the BMS is inside a quadratic shape curved tunnel and carving the surface of the Sawbone samples. (a) displacements obtained due to applying a 0.15 N.m twisting torque to the BMS head; (b) distribution of the stress along the BMS considering the applied 0.15 N.m twisting torque to the BMS head.

## III. DETAILED DESIGN AND FABRICATION OF THE BMS

The conceptual design is illustrated in Fig.2. We designed various types of BMSs and evaluated them with finite element (FE) analysis before fabrication procedure. As the first step of the design process and prior to designing the BMS geometry, we opted the nitinol (NiTi) as the screw material due to its biocompatibility for medical use and its superelasticity (i.e. providing flexibility and large deflection capability) yet greater structural strength as compared to other materials (e.g. stainless steel). The overall length, outer and core diameters of the screw depend on the size of the drilled tunnel. Hence, given the 35 mm length of the ortho-snake and 8.5 mm diameter of the drill bit (as shown in Fig. 5), we designed the BMS with a core diameter of 7.5 mm, outer diameter of 9.5 mm and overall length of 50 mm. Of note, the dimensions of ortho-snake and BMS can be adjusted based on the clinical application and decision of the surgeon. To choose type, depth, and pitch of the threads, we analyzed the traditional rigid cancellous bone screws. These screws have deep V-shape threads with larger pitch as compared to cortical bone screws. These features result in less number of threads in a given length yet greater holding capacity in cancellous bone [14]. Inspired by the design of these screws and to ensure sufficient space for the orthogonal notches along the BMS shaft, we considered V-shape threads. To make the BMS flexible, we created a through hole inside the BMS shaft (i.e. a cannulated BMS) and deep notches in the space between threads such that they intersected the through hole. The self-tapping feature was created using two symmetric sharp notches along the first three threads of the BMS next to the steering head.

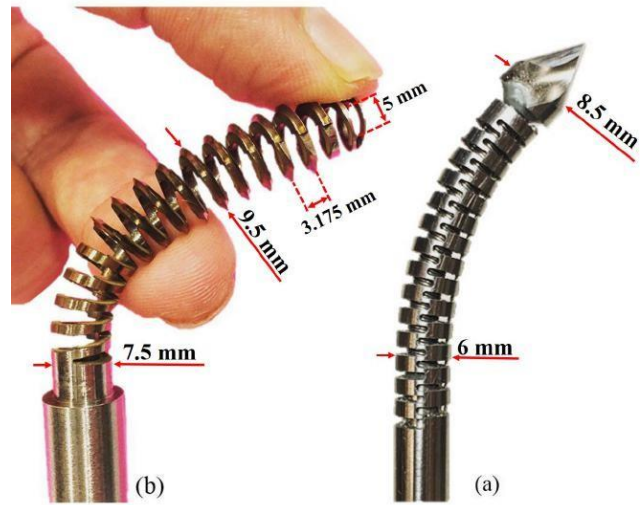

Fig. 5. (a) ortho-snake (OD=6 mm) with the flexible drill (OD=8.5 mm) passed through its 4 mm instrument channel; (b) designed and fabricated self-tapping BMS (OD=9.5 mm) for robot-assisted internal fixation of bone fractures.

To obtain the detailed design of the BMS, we performed two different FE simulations using linear tetrahedral elements in COMSOL Multiphysics® (COMSOL AB, Stockholm,

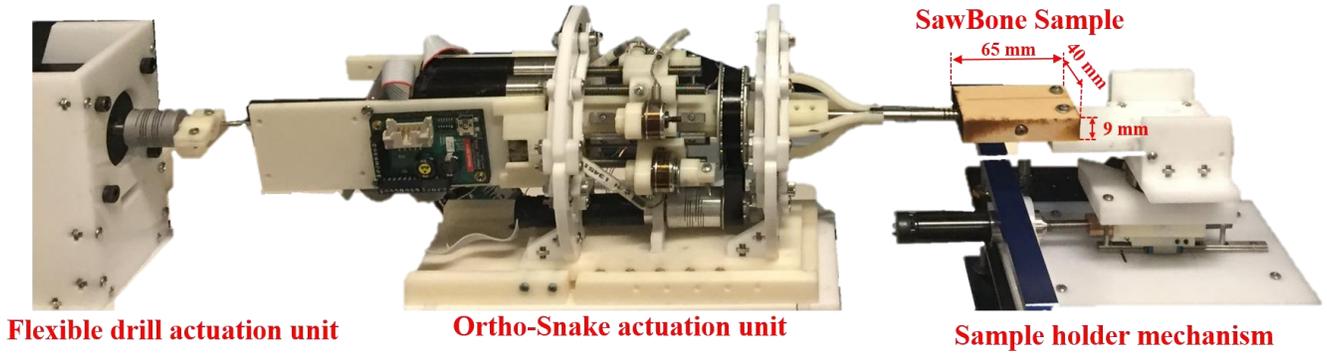

Fig. 6. Experimental setup used to perform the curved drilling experiments on the Sawbone samples.

Sweden). These simulations assessed the strength of the designed BMSs during (i) their shape adaptability, and (ii) carving the groves on the cancellous bone . Fig. 3 illustrates the FE results of the shape adaptability simulation on the finalized version of the designed cannulated BMS with inner hole diameter of 5 mm and 11 V-shape threads with 3.175 mm pitch. In this study, we assumed the threaded part of the BMS (i.e. section 2 in Fig. 2) was completely inserted into a quadratic shape curved tunnel. We then generated a FE model for the BMS. The nitinol screw elements were assumed linear, elastic and isotropic with a Young's modulus of 83 GPa and Poisson's ratio of 0.33. Based on the bending and tip displacement capability of the ortho-snake, we ap- plied a quadratic displacement boundary condition along the BMS with maximum displacement of 40 mm at its tip and minimum of zero at the beginning of the threaded section (Fig. 3a). Considering the yield strength (700 MPa) and ultimate tensile strength (1900 MPa) of nitinol, investigation of Fig. 3b reveals that the BMS is able to safely flex to the corresponding shape and does not fail/break during screw insertion into the curved tunnel.

Fig. 4 demonstrates the FE results of the case simulating the strength of the screw during carving the bone. For this particular test, we simulated an instant when the BMS has been inserted and carved the curved tunnel and the twisting torque is exerted on the screw head. Similar to the previous step, we considered linear tetrahedral elements and assumed BMS is placed inside the same quadratic shape tunnel described in the previous study while a 0.15 N.m twisting torque is applied to its head. This torque was chosen based on the reported maximum insertion torque for conventional cancellous screws with similar dimensions [14]. It is notable that this value potentially may differ from an actual BMS required insertion torque- due to the change in the stiffness, geometry, and flexibility of the screw; however, performing simulations with this initial estimation helped us to simulate more realistic situations during the design procedure. Fig. 4 demonstrates the maximum stress and displacement along the curved BMS for the aforementioned construct asserting that the BMS do not fail in this condition as well.

Fig. 5b shows the fabricated BMS from nitiniol based on the described design parameters. To fabricate this BMS, we first used a CNC machine to cut the threads, steering head, and the self tapping edges on a 12 mm nitinol rod (Kelloggs Research Lab, New Hampshire, USA). Then, we drilled a 5 mm through hole inside the threaded rod and used a wire-cut electrical discharge machine (EDM) to finalize the process and cut the notches on the surface of the BMS.

## IV. EXPERIMENTAL SETUP

To provide a consistent testing substrate during experiments, we used the 15 PCF polyurethane foam block (Sawbones; Pacific Research Laboratories, Washington, USA) as a surrogate for human cancellous bones. This synthetic bone model has been recommended by American Society for Testing and Materials (ASTM) standard F1839-08 and different researchers for testing of orthopedic implants (e.g screws [14], [16]) due to its similar values for Youngs modulus, yield strength, compressive strength, and density as cancellous bone [17]. For the experiments, we made rectangular samples (65 mm×40 mm×9 mm) from this synthetic block (Fig. 6).

Aside from ortho-snake and the flexible drill bit, as illustrated in Fig. 6, the robotic system includes three main subsystems: (i) ortho-snake actuation unit; (ii) flexible drill actuation unit; and (iii) a two degrees-of- freedom sample holder mechanism. The ortho-snake's actuation unit has 4 motors (RE16, Maxon Motor Inc.) with spindle drives (GP16, Maxon Motor, Inc.) to pull the ortho-snake's cables. It also has four load cells (Model 31Mid, Honeywell Inc.) to read each of the cable tensions. The tool actuation unit has one stepper motor (DMX-UMD-23, Arcus Technology, Inc.) to rotate the flexible drill. The sample holding mechanism consists of a brushless DC motor (RE16, Maxon Motor, Inc.) with a spindle drive (GP16, Maxon Motor, Inc.) that moves a cart carrying the sample on a linear stage toward the flexible drill integrated with the ortho-snake [12]. A custom C++ interface performs independent velocity or position control of each motor of the ortho-snake's actuation unit using libraries provided by the company.

## V. EXPERIMENTS AND RESULTS

The following sections describe the experiments and results of evaluating the performance of the proposed robot-assisted internal fixation procedure.

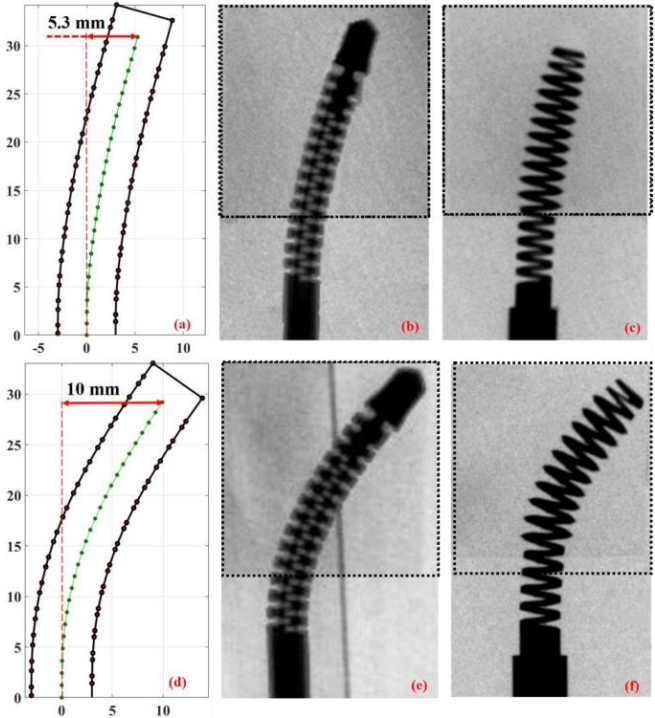

Fig. 7. First row is the results of the curved drilling experiment with 5 N cable pulling tension: (a) registered ortho-snake shape; (b) X-ray image of the integrated ortho-snake and the cutting tool inside the drilled Sawbone sample; and (c) X-ray image of the inserted BMS into the drilled curved tunnel. Second row is the results of the curved drilling experiment with 25 N cable pulling tension: (d) registered ortho-snake shape; (e) X-ray image of the integrated ortho-snake and the cutting tool inside the drilled Sawbone sample; and (f) X-ray image of the inserted BMS into the drilled curved tunnel.

### A. Investigation of the Shape-Adaptability and Tapping-Capability of the BMS in the Drilled Curved Tunnels

To evaluate the proposed RAOP for internal fixation of bone fractures, we first drilled a curved tunnel in a synthetic bone sample using the described robotic system and then inserted the fabricated BMS. To accomplish this, we first fixed a synthetic bone sample on the sample holder, and inserted the flexible drill into the ortho-snake and cutting unit. Then, in order to check various drilling possibility as well as the shape adaptability, carving capability, and strength of the BMS in the drilled tunnels, we considered three different cable tensions (i.e. 5, 10, and 25 N). The maximum cable tension was limited by the maximum load capacity of the utilized stainless steel braided cable (8930T18, McMaster-Carr). Of note, higher cable tension results in a higher bend in the ortho-snake, which subsequently imposes a higher displacements at the BMS tip. For each experiment, we simultaneously controlled the cable tension, sample holder velocity (0.15 mm/s), and rotation speed of the drill (2250 rpm). In [13], we showed that with this rotational speed and feeding velocity of sample, the ortho-snake will not buckle and the drilling time would be minimum. As shown in Fig. 7b and Fig. 7e, after each experiment, to measure maximum displacement of the ortho-snake tip and subsequently the BMS, we took an X-ray image of the integrated flexible drill with the ortho-snake inside the drilled sample using a C-arm machine (ARCADIS Orbic, Seiemens; Munich, Germany). A 2D-3D registration method was implemented on these images to register the ortho-snake shape and calculate its deflection [18]. Fig. 7a and Fig. 7d show the results of registration and the corresponding drilled tunnel for the conducted experiments with 5 N and 25 N cable tensions. As shown, the maximum displacement (i.e. 10 mm) corresponds to the 25 N cable tension while a 5.3 mm displacement happened due to a 5 N cable tension. Then, to check the self-tapping capability and shape adaptability of the BMS, as shown in Fig. 8, we tapped the drilled samples using BMS. This figure shows the cross section of the tapped samples along the curved tunnel drilled with 25 N and 10 N cable tensions. We fastened and unfastened the BMS into each sample to ensure repeatability and strength of the screw. Fig. 7e and Fig. 7f demonstrate the X-ray images of the BMS fastened through the tapped curved tunnels in bone samples with two different curvatures. The provided media file shows the X-ray sequences of shape adaptability of the BMS during its insertion through the tapped samples shown in Fig. 7.

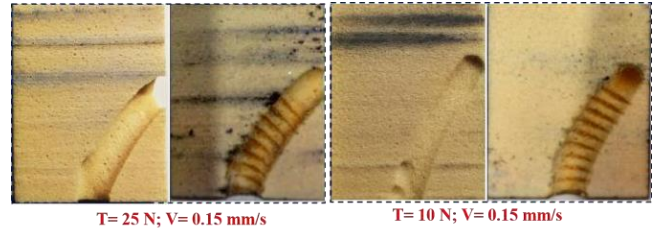

Fig. 8. Cross sections of the drilled and tapped curved tunnels with two different cable tensions demonstrating the self-tapping capability of the BMS.

### B. Investigation of the BMS fixation Strength

We also performed a preliminary experiment to evaluate the mechanical strength of a simulated fracture with and without using the BMS similar to the performed experi- ments in [5]. As shown in Fig. 9, to simulate a potential bone fracture, we arbitrarily made a V-shape transversal cut passing the center of the curved tunnel with 2 mm width. To investigate the mechanical strength created with the BMS fixation, we added an eccentric variable load with respect to the BMS entry point (i.e. 15 mm offset) causing a bending moment on the screw and, therefore, having an adverse effect on the fixated fracture. We then fixated the fracture by passing the screw through a curved tunnel drilled with 25 N cable tension (as shown in Fig. 7e and Fig. 8). We increased the magnitude of the eccentric load until the failure happens. The maximum eccentric load for this particular loading condition and drilled tunnel obtained as 155 N, when the Sawbone sample could not withstand this loading and broke. We repeated the same experiment without fixating the simulated fracture with the BMS and obtained 10 N maximum eccentric load capacity.

## VI. DISCUSSION AND CONCLUSION

In this paper, we introduced a novel robot-assisted orthopedic procedure for internal fixation of bone fractures. Uti-lizing a custom designed continuum manipulator and flexible cutting instruments, we proposed a curved drilling technique together with a novel bendable medical screw to stabilize the fractured bone by creating adequate compression. This novel treatment approach has the potential to provide clinicians with freedom to choose the entry point location as well as the steering capability during drilling the bone.

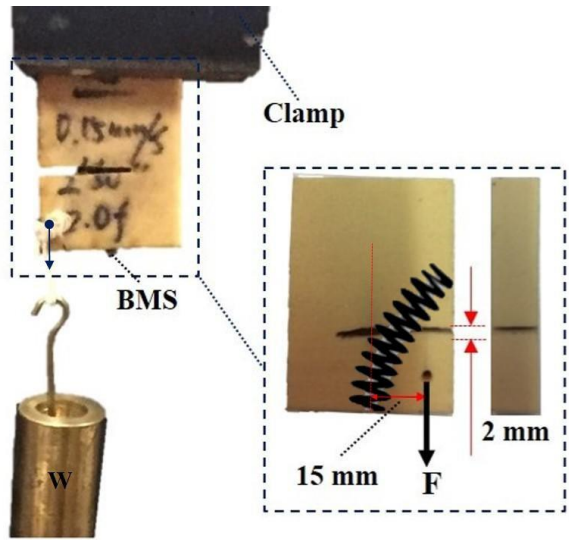

Fig. 9. Experimental setup designed to evaluate the BMS fixation strength on a Sawbone sample with a simulated transversal fracture drilled with 25 N cable tension and experiencing an eccentric variable load. The maximum eccentric load for this experiment obtained as $W \approx 15.5$ kg.

We used the robotic system to create various C- shape curved tunnels and successfully evaluated the shape-adaptability and self-tapping capability of the manufactured bendable screw. Considering the results obtained with these tests, the fabricated BMS demonstrated its safe morphability (during and after tapping procedure) without failure/break inside drilled tunnels of various curvatures. Of note, this important feature together with the steerability of the ortho-snake has the potential to enable clinicians to plan curved tunnels that provide maximum stability and compression for the fractured site. In future, we will utilize the insight obtained in this study to plan and control the ortho-snake to drill curved tunnels for internal fixation of realistic fractures that provides maximum stability. Further, additional studies are needed to evaluate the performance of the BMS when the path includes multi-curvatures (e.g. an S-bend shape).

Our preliminary experiments demonstrated the satisfactory performance of the robotic system for fracture fixation in a basic scenario. The main goal of this experiments was to acquire a preliminary understanding about the weight bearing capacity of the proposed BMS. In the future, additional experiments are needed to verify the fracture reduction capability of our proposed method for more diverse situations including geometry of the fractured bones, location of the eccentric load, curvature of the drilled tunnel and etc. Extensions to this work may also include biomechanical studies to investigate the pros and cons of the proposed internal fixation procedure with the conventional approaches through performing finite element simulations and cadaver studies.